\documentclass[10pt,twocolumn,letterpaper]{article}

\usepackage{wacv}              

\usepackage{graphicx}
\usepackage{booktabs}
\usepackage{gensymb}

\usepackage{times}
\usepackage{epsfig}
\usepackage{amsmath}
\usepackage{amssymb}
\usepackage{multirow}
\usepackage{subcaption}
\DeclareMathOperator*{\argmin}{arg\,min}

\usepackage[pagebackref,breaklinks,colorlinks]{hyperref}

\usepackage[capitalize]{cleveref}
\crefname{section}{Sec.}{Secs.}
\Crefname{section}{Section}{Sections}
\Crefname{table}{Table}{Tables}
\crefname{table}{Tab.}{Tabs.}

\def\wacvPaperID{987} 
\def\confName{WACV}
\def\confYear{2024}

\begin{document}

\title{Face Presentation Attack Detection by Excavating Causal Clues and Adapting Embedding Statistics}
\vspace{-5mm}

\author{Meiling Fang$^{1}$, Naser Damer$^{1,2}$ \\ 
$^{1}$Fraunhofer Institute for Computer Graphics Research IGD,
Darmstadt, Germany\\
$^{2}$Department of Computer Science, TU Darmstadt,
Darmstadt, Germany\\
{\tt\small Email: {meiling.fang@igd.fraunhofer.de}}
}

\maketitle

\begin{abstract}
\vspace{-3mm}
Recent face presentation attack detection (PAD) leverages domain adaptation (DA) and domain generalization (DG) techniques to address performance degradation on unknown domains. However, DA-based PAD methods require access to unlabeled target data, while most DG-based PAD solutions rely on a priori, i.e., known domain labels. Moreover, most DA-/DG-based methods are computationally intensive, demanding complex model architectures and/or multi-stage training processes. This paper proposes to model face PAD as a compound DG task from a causal perspective, linking it to model optimization. We excavate the causal factors hidden in the high-level representation via counterfactual intervention. Moreover, we introduce a class-guided MixStyle to enrich feature-level data distribution within classes instead of focusing on domain information. Both class-guided MixStyle and counterfactual intervention components introduce no extra trainable parameters and negligible computational resources. Extensive cross-dataset and analytic experiments demonstrate the effectiveness and efficiency of our method compared to state-of-the-art PADs. The implementation and the trained weights are publicly available\footnote{\url{https://github.com/meilfang/CF-PAD}}.
\vspace{-3mm}
\end{abstract}

\begin{figure}[ht!]
\begin{center}
\includegraphics[width=0.88\linewidth]{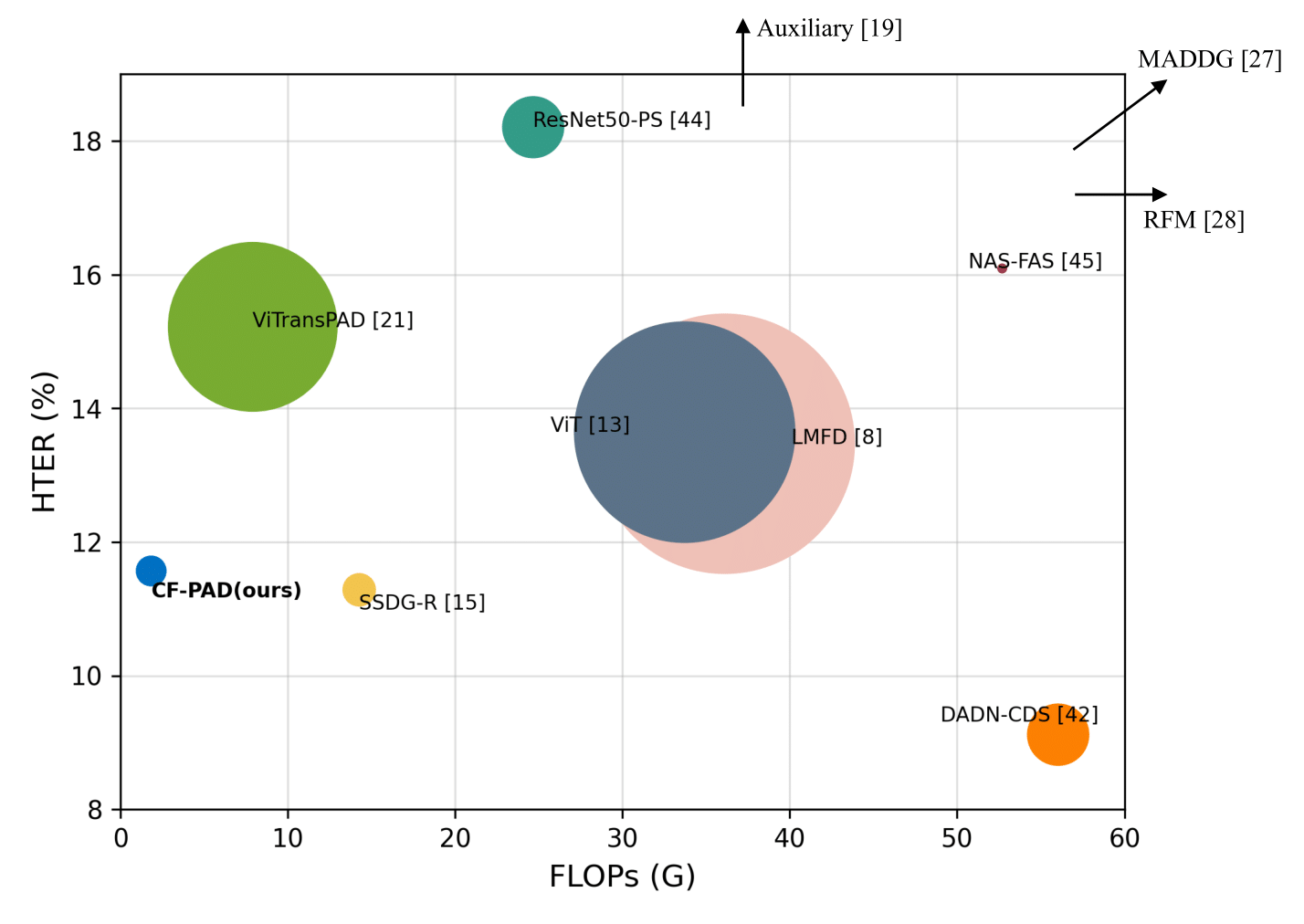}
\vspace{-2mm}
\caption{PAD performance vs. Computational complexity. x-axis indicates the complexity of the model by FLOPs, y-axis indicates the PAD performance represented by HTER (\%), and the radius of the circle indicates the number of trainable parameters. An ideal PAD tends to be located in the bottom left corner with a small circle, as achieved by our CF-PAD solution (in blue).}
\vspace{-9mm}
\label{fig:computation_complexity}
\end{center}
\end{figure}

\vspace{-3mm}
\section{Introduction} 
Face recognition \cite{DBLP:conf/cvpr/DengGXZ19,DBLP:conf/cvpr/BoutrosDKK22,BOUTROS2023104688} has become a remarkable technique for identity authentication and has been widely deployed in our daily lives. However, face recognition is vulnerable to presentation attacks (PAs) \cite{casia_fas,oulu_npu,DBLP:conf/eccv/ZhangYLYYSL20,DBLP:journals/pr/FangDKK22}, such as printed images, replayed videos, and 3D masks. Therefore, face presentation attack detection (PAD) plays an important role in securing face recognition from such PAs. Despite the promising performance in intra-dataset evaluation by deep learning-based face PAD techniques \cite{DBLP:journals/tbbis/YuLSXZ21,DBLP:journals/pami/YuWQLLZ21,DBLP:conf/icb/FangAKD22}, the PAD performance significantly degrades under the  more realistic cross-domain scenario. 

Recently, a variety of domain adaptation (DA) and domain generalization (DG) based PAD methods \cite{DBLP:journals/tcsv/YanZH22,DBLP:journals/tbbis/WangWDG22,DBLP:conf/cvpr/ShaoLLY19,DBLP:conf/cvpr/JiaZSC20,DBLP:conf/aaai/ShaoLY20} are proposed to boost the PAD generalizability. DA-based PAD methods \cite{DBLP:journals/tifs/LiLCWHK18,DBLP:conf/icb/WangHSC19} learn a discriminative feature space by accessing the labeled source domains and unlabeled target domains. However, this target data is typically unavailable in real-world scenarios.
In contrast with DA-based methods leveraging the target testing data, DG-based PAD methods \cite{DBLP:conf/cvpr/ShaoLLY19,DBLP:conf/cvpr/JiaZSC20,DBLP:conf/aaai/ShaoLY20,DBLP:conf/aaai/ChenYSDTLHJ21} aim to generalize the knowledge obtained from multiple source datasets to the unseen target domain and has achieved remarkable progress. However, these efforts only attempt to alleviate the problems caused by different data distributions by modelling the statistical dependencies between inputs and outputs. They do not excavate the intrinsic causal mechanisms. 
In addition, DG-based PAD methods rely on a priori knowledge that the domain label of each source dataset is known. This is challenging to be satisfied in practical situations and it is hard to define the representation of a \textit{domain} well enough. Moreover, most existing DG-based face PAD methods are computationally intensive due to complex model architectures \cite{DBLP:conf/eccv/HuangSLCXYAY22,DBLP:journals/tcsv/YanZH22} or multi-stage/multi-network training \cite{DBLP:conf/icmcs/LiuCDLZX22,DBLP:conf/aaai/ChenYSDTLHJ21}.

To address the above issues, this work presents a structural causal model (SCM) to formulate the face PAD problem, aiming to uncover the causality hidden in the high-level representative features (as shown in Fig. \ref{fig:scm}) and enhance the PAD generalizability, without requiring complex architectures. The contributions of our work are as follows:
1) \textcolor{black}{We model the compound DG-based PAD from a causality-based view linking it to the model optimization. Our method relaxes the need for a priori domain knowledge compared to most DA/DG based PADs, which rely on domain labels or target data.}
2) we propose to boost the PAD generalizability by enriching the feature-level training data distribution via our proposed class-guided MixStyle and separating the causal factors from high-level representation via counterfactual interventions. Both components require no extra trainable parameters and very little computation complexity, as illustrated in Fig. \ref{fig:computation_complexity}.
3) We conduct extensive experiments under various cross-dataset scenarios and the analytical results demonstrate the effectiveness and efficiency of our method in comparison to existing works.

\vspace{-2mm}
\section{Related Work} 

\paragraph{Face Presentation Attack Detection} 
Many face PAD solutions \cite{Fang_2022_WACV,DBLP:journals/pami/YuWQLLZ21,DBLP:journals/tbbis/YuLSXZ21,DBLP:conf/cvpr/LiuJ018,DBLP:conf/fgr/FangBKD21,DBLP:conf/bmvc/DamerD16} have been proposed and shown good performance with the recent advances of deep-learning techniques, especially in intra-dataset evaluations. Recently, a number of PAD works \cite{DBLP:journals/tcsv/YanZH22,DBLP:journals/tbbis/WangWDG22,DBLP:conf/cvpr/ShaoLLY19,DBLP:conf/cvpr/JiaZSC20,DBLP:conf/aaai/ShaoLY20} turned to leverage DA and DG techniques to target the 
challenging cross-domain scenario.
DA transfers the knowledge from the source domain to the target domain, where the unlabeled target data is accessed in the training process. DA-based face PAD methods align the feature space between source and target domains by minimizing the maximum mean discrepancy (MMD) \cite{DBLP:journals/tifs/LiLCWHK18} and adversarial training \cite{DBLP:conf/icb/WangHSC19}.
However, collecting unlabeled target domain data is very difficult and laborious. Moreover, in real-world scenarios, there is usually no information available for the target domain during training.
In comparison with DA-based methods, DG-based PAD methods are more practical for real-world deployment because they do not require the acquisition of target data.
DG-based PAD methods learn a model from multiple datasets to obtain a generalized feature representation that allows generalizing well to the unseen target domain. There are various types of DG-based PAD methods: 1) adversarial training \cite{DBLP:conf/cvpr/ShaoLLY19,DBLP:conf/cvpr/JiaZSC20}, 2) meta-learning \cite{DBLP:conf/aaai/ShaoLY20,DBLP:conf/aaai/ChenYSDTLHJ21}, simulating the domain shift by dividing source data into meta-train and meta-test sets. 
Most existing DA- or DG-based PAD solutions are computationally intensive because of the use of large model architectures such as ViT \cite{DBLP:conf/eccv/HuangSLCXYAY22}, DADN-CDS \cite{DBLP:journals/tcsv/YanZH22} and MADDG \cite{DBLP:conf/cvpr/ShaoLLY19} consisting of multiple feature extractors and discriminators.
Different from the above-mentioned face PAD methods, we approach the face PAD from a causal perspective and focus on learning the causal representations by analyzing the contribution of features to PAD decisions. Our method is an efficient model that boosts the generalizability of a simple single network for face PAD by incorporating two components. These two components require no extra parameters during training and thus eliminating the need for multiple networks and/or computationally costly components (e.g., attention blocks). In addition, these two components are removed during the inference.
\vspace{-4mm}
\paragraph{Causal Inference}
Causal inference \cite{DBLP:journals/jmlr/Pearl10,DBLP:journals/ai/FentonNC20}, analyzing the correlation between the variables and the final results, has gained increasing attention in recent years. The causal inference has been successfully applied in many computer vision tasks to enhance the generalizability of models.
For example, Rao \etal \cite{DBLP:conf/iccv/Rao0L021} proposed to learn the attention with counterfactual causality by measuring the attention quality and encouraging the model to generate high-quality attention for fine-grained image recognition. Chen \etal \cite{DBLP:conf/iccv/0002LL021} proposed a counterfactual analysis approach for human trajectory prediction. They investigated the causality between the predicted trajectories and inputs and mitigated the negative effect brought by the environment bias. 

A recent work, Causal Intervention for Face Anti-spoofing (CIFAS) method \cite{DBLP:conf/icmcs/LiuCDLZX22}, was proposed to target the generalizability of face PAD. 
CIFAS \cite{DBLP:conf/icmcs/LiuCDLZX22}, \textcolor{black}{as DG PAD solution}, was facilitated based on a priori that domain labels are known and assumed that data from various domains effecting the PAD performance. Therefore, the PAD performance was improved by the preconceived domain-specific features in CIFAS, \textcolor{black}{i.e., the pre-learned domain features are used for backdoor adjustment.} In contrast, our goal is to automatically discover the intrinsic causality from the learned high-level features without any priori knowledge, \textcolor{black}{i.e. as a compound DG PAD task}. Moreover, CIFAS \cite{DBLP:conf/icmcs/LiuCDLZX22} is a three-stage face PAD framework including two networks and an additional domain feature extraction process. Compared to the computationally intensive CIFAS, our proposed solution is an end-to-end model containing two components without additional trainable parameters other than a lightweight feature extractor and a classifier.
 

\vspace{-2mm}
\section{Methodology} 
\label{sec:method}
\vspace{-3mm}
\begin{figure}[ht!]
\begin{center}
\includegraphics[width=0.5\linewidth]{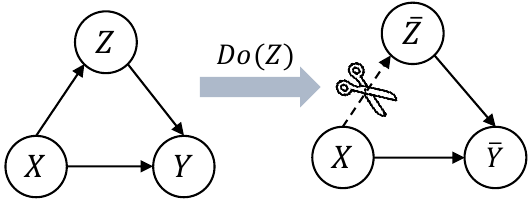}
\vspace{-2mm}
\caption{SCM with variables $X$ (face input), $Y$ (prediction), and $Z$ (high-level feature), representing the PAD task $(X,Z) \rightarrow Y$. Performing CI on feature $Z$, cutting the link between $X \rightarrow Z$, can force the $Z$ independent of its causal parent $X$. The detailed CI is presented in Sec. \ref{ssec:CI}.}
\label{fig:scm}
\end{center}
\vspace{-8mm}
\end{figure}

\subsection{PAD in the view of contextual reasoning}
\label{ssec:background}
\begin{figure*}[thbp!]
\begin{center}
\includegraphics[width=0.97\linewidth]{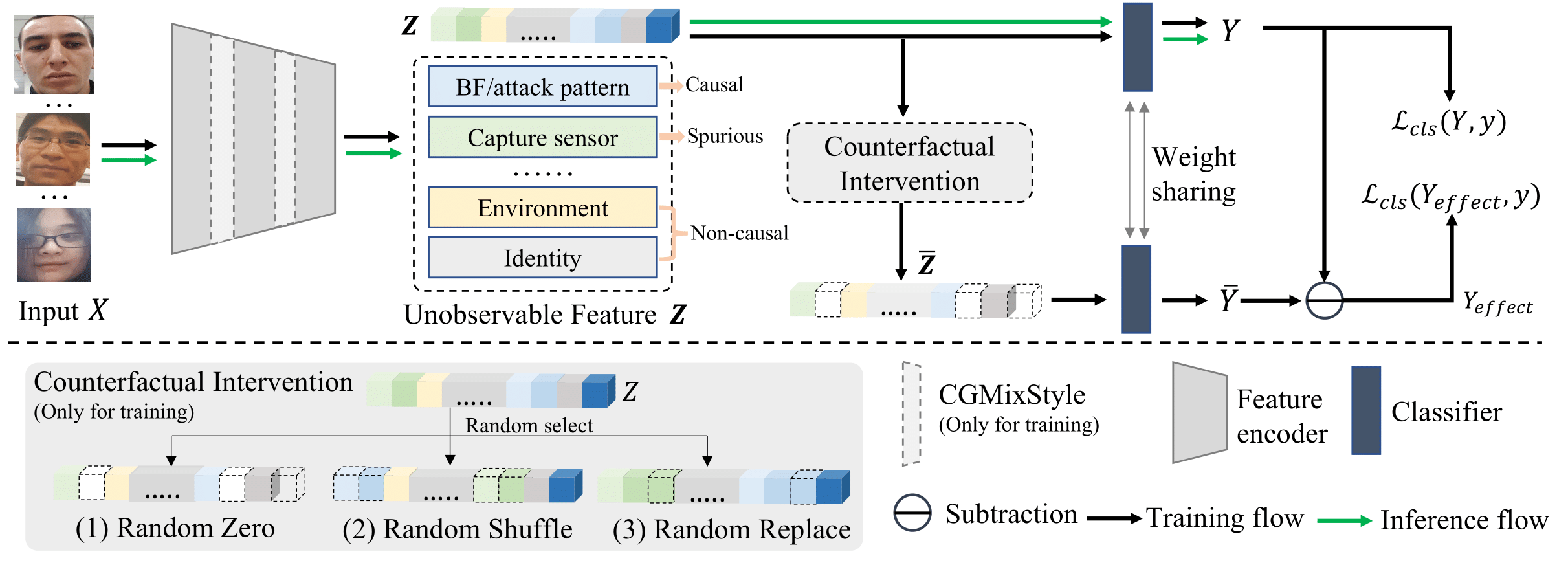}
\vspace{-2mm}
\caption{The workflow of the proposed CF-PAD solution. During the training process, CGMixStyle enriches the diversity of feature-level training data distribution and three CIs guides the model to uncover the causal factors that truly contributes to the PAD decision. Both CGMixStyle and CI are removed in the inference phase.}
\label{fig:workflow}
\end{center}
\vspace{-7mm}
\end{figure*}

We provide a background on causal inference \cite{DBLP:journals/jmlr/Pearl10,DBLP:journals/ai/FentonNC20} and model optimization \cite{DBLP:conf/nips/Vapnik91} that will form the basis for understanding the rest of this paper. 

We first introduce the \textbf{SCM} $\mathcal{M}$ to help formulate the underlying causal structure of the PAD task. A convention SCM consists of a set of causal variables and causal links. As shown in Fig.  \ref{fig:scm}, $X$ is the input (i.e. face images), $Y$ is the final prediction (i.e. bona fide or attack), and $Z$ is the intermediate high-level features. 
The PAD task here is to predict $Y$ from $X$, relying on the high-level features $Z$ hidden in the inputs, presented as $(X,Z) \rightarrow Y$. The PAD models are commonly optimized by supervising the final prediction $Y$, overlooking the effect of $Z$ on $Y$. Causal inference \cite{DBLP:journals/ai/FentonNC20}, the analysis of causal correlation between these variables, facilitates thinking outside the box.
In practice, we can analyze the causalities by manipulating the values of variables and observing the caused effect. This manipulation is termed \textit{intervention} in causal inference \cite{DBLP:journals/ai/FentonNC20} and represented by an idiomatic symbol $do(\cdot)$. 


Next, we link the intervention with the model optimization by recalling the conventional empirical risk minimization (ERM) \cite{DBLP:conf/nips/Vapnik91}. 
Assume a model $h$, accessing training data from a distribution $p(X,Y) \in \mathcal{P}$, is trained to minimize the empirical risk $\hat{R}$:
\begin{equation}
\begin{aligned}
    \hat{R}_{p(X,Y)}(h) = \hat{\mathbb{E}}_{p(X,Y)}[\mathcal{L}(Y, h(X))]
\end{aligned}   
\end{equation}
where $\mathcal{L}$ refers to loss function. The objective of intervention risk minimization associating to an intervention $I$ is:
\begin{equation}
\begin{aligned}
    \hat{R}_{p_{(X, Y |I)}}^{\mathcal{M}} (h) = \hat{\mathbb{E}}_{p_{(X, Y |I)}}^{\mathcal{M}}[\mathcal{L}(Y, h(X))] 
\end{aligned}   
\end{equation}
where $p_{(X, Y |I)}$ indicates the intervened data distribution.
When a set of intervention operations $I \in \mathcal{I}$ are performed over the SCM, the worst-case intervened data distribution is:
\begin{equation}
\begin{aligned}
    \hat{R}_{p_{(X, Y |I)}}^{\mathcal{M}} (h) = \max_{I \in \mathcal{I}} \, \hat{\mathbb{E}}_{p_{(X, Y |I)}}^{M}[\mathcal{L}(Y, h(X))] \\
\end{aligned}   
\end{equation}
Finally, a generalized model can be obtained by solving
\begin{equation}
\begin{aligned}
    h^* = \argmin_{h} \, \max_{I \in \mathcal{I}} \,\hat{\mathbb{E}}_{p_{(X, Y |I)}}^{\mathcal{M}}[\mathcal{L}(Y, h(X))]
\end{aligned}
\label{eq:erm_scm}
\end{equation}

As the min-max optimization problem is very challenging, Equation \ref{eq:erm_scm} can be approximated by 1) enriching the diversity of the training data distribution and 2) increasing the diversity of interventions \cite{DBLP:conf/iccv/ZhangWWLKLG21}. 
This theoretical basis motivates us to develop a generalized PAD by solving this approximation problem from both perspectives.
The pipeline of the proposed CF-PAD solution is shown in Fig. \ref{fig:workflow}, containing the two components we proposed in this work, class-guided MixStyle (CGMixStyle) and Counterfactual Intervention (CI).
CGMixStyle aims to enrich the diversity of training data (on the feature-level), while CI mines causal related features by performing a set of counterfactual interventions. The details are presented in Sec. \ref{ssec:CGMixStyle} and \ref{ssec:CI}, respectively.

\begin{figure}[th!]
\begin{center}
\includegraphics[width=0.98\linewidth]{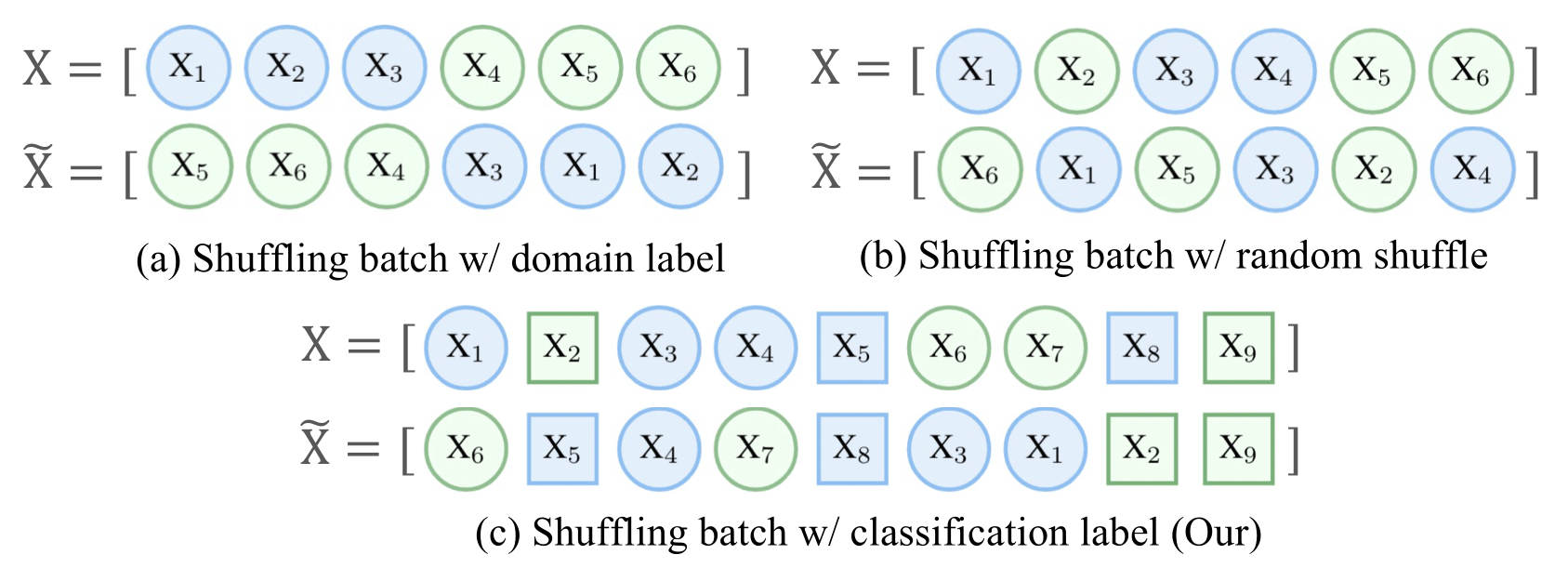}
\vspace{-2mm}
\caption{The comparison of shuffling operation between MixStyle (a,b) and our CGMixStyle (c). Color and shape refer to domain and class, respectively. Class-guided shuffle restrict the shuffling within the same class (bona fide or attack in PAD) without the prior domain information.}
\label{fig:shuffle_operation}
\end{center}
\vspace{-6mm}
\end{figure}

\vspace{-1mm}
\subsection{Class-guided MixStyle}
\label{ssec:CGMixStyle}
\vspace{-1mm}
Due to the privacy concerns and the labor-intensive nature of data collection, most existing face PAD datasets are limited in identity, capture environment, and attack type. Therefore, increasing the training data diversity by incorporating more training data is the less realistic option. Inspired by the recently introduced MixStyle \cite{DBLP:conf/iclr/ZhouY0X21}, we propose a class-guided MixStyle to increase the distribution of the available limited training data by mixing the feature statistics within the same class. 
The MixStyle \cite{DBLP:conf/iclr/ZhouY0X21} is formulated as the following: 
\vspace{-2mm}
\begin{equation}
\small
\label{eq:feature_calculatio}
\begin{aligned}
    \gamma = \lambda \sigma(x) + (1-\lambda) \sigma(\tilde{x}) 
    \\
    \beta = \lambda \mu(x) + (1-\lambda) \mu(\tilde{x}) 
\end{aligned}   
\end{equation}
Where $x \in \mathbb{R}^{B \times C \times H \times W}$ refers to the input batch feature and $\tilde{x}$ indicates the corresponding reference batch from $x$. $\sigma(.), \mu(.) \in \mathbb{R}^{B \times C}$ are mean and standard deviation computed across the spatial dimension within each channel of each sample. $\lambda \in \mathbb{R}^B$ are weights sampled from the Beta distribution. 
The final mixed feature statistic is applied to the styled normalized $x$ as:
\vspace{-3mm}
\begin{equation}
\small
\label{eq:mixstyle}
    MixStyle(x) = \gamma \frac{x-\mu(x)}{\sigma(x)} + \beta
\end{equation}
\vspace{-3mm}

In MixStyle \cite{DBLP:conf/iclr/ZhouY0X21}, two shuffle operations are considered to generate $\tilde{x}$ from $x$, domain label guided and random shuffle, as shown in Fig.  \ref{fig:shuffle_operation} (a) and (b). After shuffling, data can be produced in new stylized domains by following the equations \ref{eq:feature_calculatio} and \ref{eq:mixstyle}. Meanwhile, the discriminative features, such as outline and shape, for general computer vision classification task, is unchanged.
However, face PAD relies on the subtle intrinsic attack clues that are assumed to be part of the image style. 
Therefore, both shuffle operations pose a potential risk of degradation in PAD performance, as the style transfer may be performed between bona fide and attack samples. 
Consequently, we propose to modify the MixStyle concept by restricting the shuffle operation within the same class, as illustrated in Fig. \ref{fig:shuffle_operation} (c). Specifically, in Eq. \ref{eq:feature_calculatio}, each sample in $\bar{x}$, generated by class-guided shuffle, belongs to the same class as each sample in $x$.
This "class-guided" MixStyle is designed to produce more diverse features without a confusion of bona fide and attack features, and thus resulting in a better generalized representations. The effectiveness of class-guided shuffling is demonstrated in later in Tab.  \ref{tab:shuffle_operations}.

\vspace{-2mm}
\subsection{Counterfactual Intervention} 
\label{ssec:CI}

In an optimal situation, a well-generalized PAD should be independent of some factors, such as sensor type, environment, and identity, as represented by the dashed rectangular box in Fig. \ref{fig:workflow}. However, PAD solutions suffer from the low generalization on unknown data \cite{DBLP:conf/icb/PurnapatraSBDYM21,DBLP:conf/eccv/ZhangYLYYSL20,DBLP:conf/eccv/XuLNX14,DBLP:conf/cvpr/FangHD23}.
Therefore, we build an assumption regarding the high-level feature $Z$ that each feature, representing for input images $X$, is constructed from a mix of different factors: 1) The class-dependent information encoded in $Z$ is causal factors, associated with a universal and invariant patterns of bona fide or attack. 2) The information independent of class in $Z$ is considered as non-causal factors, such as identity information and capture environments. 3) Capture sensors is considered to be the spurious factor as several existing cross-domain PADs \cite{DBLP:conf/eccv/ZhangYLYYSL20,DBLP:journals/tcsv/YanZH22,DBLP:conf/icmcs/LiuCDLZX22} stated that the PAD performance is affected by the capture devices.

Our goal is to identify the factors in $Z$ that truly contribute to the class $Y$. 
As mentioned in Sec. \ref{ssec:background} and illustrated in Fig.  \ref{fig:scm}, we can perform intervention on the learned feature $Z$ to measure its contribution and encourage model to produce more generalized features. Specifically, We propose to adopt \textit{counterfactual intervention} \cite{hagmayer2007causal,Tyler2015} to analyse the effect of high-level features.
Counterfactual intervention ($do(Z=\bar{Z})$) refers to generate a counterfactual sample where the causal factors are removed. 
However, generating (true) counterfactual features is impractical, because the causal factors in high-level feature is unobservable and unformulated.

In our work, we present three simple interventions, random zero, random shuffle, and random replace, to encourage the model to uncover the causal factors from the confounding feature $Z$. This uncovering is facilitated by comparing the effect of the original features $Z$ and counterfactual features $\bar{Z}$, as shown in Fig.  \ref{fig:workflow}. The causal effect is then formulated based on the difference of the original prediction and its counterfactual alternative prediction:
\vspace{-2mm}
\begin{equation}
    Y_{effect} = Y(Z) - Y(\bar{Z}) 
\end{equation}
where $Y(\bar{Z}) = f(do(Z = \bar{Z}), X=X)$ and $f(\cdot)$ is the final classifier, catching the changes in final decision caused by the interventions. The loss function is formulated as:
\vspace{-2mm}
\begin{equation}
    \mathcal{L} = \mathcal{L}(Y, y) + \lambda \mathcal{L}(Y_{effect}, y)
\end{equation}
where $y$ is the ground-truth label and $\lambda$ is the loss weight. $\mathcal{L}$ is the cross-entropy loss in our case. The former loss optimizes the model with the actual correct predictions and the latter loss penalizes the \textit{wrong} results based on the counterfactual samples, and thus guiding the model to mine the casual factors. 

\subsection{Enabling computational efficiency}
During the training phase, CGMixStyle and CI both have no extra trainable parameters. 
In the case of CGMixStyle, the gradients of $\mu$ and $\sigma$ in Eq. \ref{eq:feature_calculatio} are blocked in the computation graph, i.e. no back-propagation. 
In the case of CI, the introduced extra computational cost is an additional feed-forward path of the counterfactual features to the final classifier, which shares the weight of the original features, i.e., no extra gradient computation. 
Moreover, during the inference phase, CGMixStyle and CI are both removed, as the green flow shown in Fig. \ref{fig:workflow}. 
Therefore, the whole computation cost of CF-PAD is very low (whether for training or inference) and minimalistic when compared to other DA- and DG-based PAD framework which requires the complicated model architectures \cite{DBLP:conf/eccv/HuangSLCXYAY22,DBLP:journals/tcsv/YanZH22,DBLP:conf/cvpr/ShaoLLY19}, multi-stage training process \cite{DBLP:conf/icmcs/LiuCDLZX22}, or iterative meta-learning steps \cite{DBLP:conf/aaai/ShaoLY20,DBLP:conf/aaai/ChenYSDTLHJ21}. 
The comparison results are discussed later in Sec. \ref{ssec:complexity} and illustrated in Fig. \ref{fig:computation_complexity}.

\vspace{-3mm}
\section{Experiments} 
\subsection{Datasets}

Following the \cite{DBLP:conf/icmcs/LiuCDLZX22,Fang_2022_WACV,DBLP:conf/cvpr/LiPWK18,DBLP:conf/cvpr/ShaoLLY19,DBLP:conf/aaai/ShaoLY20,DBLP:conf/fgr/PanwarSSPG21}, we conduct experiments on five publicly available face PAD datasets: MSU-MFSD \cite{msu_mfs} (denoted as M), CASIA-FASD \cite{casia_fas} (denoted as C), Idiap Replay-Attack \cite{replay_attack} (denoted as I), OULU-NPU \cite{oulu_npu} (denoted as O), and CelebA-Spoof \cite{DBLP:conf/eccv/ZhangYLYYSL20} (denoted as CA).

The \textbf{MSU-MFSD} \cite{msu_mfs} dataset is comprised of 440 videos across from 35 subjects and contains two types of attacks, printed photo attacks and replay attacks.
The \textbf{CASIA-FASD} \cite{casia_fas} dataset consists of 600 videos from 50 subjects and includes three types of attacks: warped photo attack, cut photo attack, and video replay attack.
The \textbf{Idiap Replay-Attack} dataset \cite{replay_attack} contains 300 videos from 50 subjects and includes two attack types: print attacks and replay attacks.
The \textbf{OULU-NPU} \cite{oulu_npu} is a mobile face PAD dataset, consisting of 5940 videos from 55 subjects using six different mobile phones. 
The \textbf{CelebA-Spoof} \cite{DBLP:conf/eccv/ZhangYLYYSL20} is a large-scale face PAD dataset collecting from the web, comprising 625,537 images from 10,177 subjects. The dataset is diverse in terms of subjects, illumination, sensors, and attack types and contains print, replay, 3D mask, and paper cut attacks.

All the experiments are cross-dataset scenarios and can be categorized into two groups based on the number of training datasets: multi-source , limited-source, single-source scenarios.
Multi-source scenario contains five cases (training dataset(s) → testing dataset) : O\&C\&I → M, O\&M\&I → C, O\&C\&M → I, I\&C\&M → O, O\&C\&M → CA, and limited-source scenarios contains M\&I → C and M\&I → O cases by following works \cite{DBLP:conf/aaai/ChenYSDTLHJ21,DBLP:conf/icmcs/LiuCDLZX22,DBLP:conf/cvpr/JiaZSC20,DBLP:conf/cvpr/Wang0SC20}. The first four cases are widely-used evaluation protocols for cross-domain PAD \cite{DBLP:journals/tbbis/YuLSXZ21,DBLP:journals/pami/YuWQLLZ21,Fang_2022_WACV,DBLP:conf/icip/MingYAVLB22} and thus utilized for the following analytical experiments in Sec. \ref{ssec:ablation_studies}.
Single-source scenario includes 12 cases by all the combinations of selecting one dataset from O,C,I,and M as the training data and testing separately on the other three datasets, by following the protocol in \cite{DBLP:journals/tifs/WangHSC21,DBLP:conf/cvpr/Wang0SC20,DBLP:conf/aaai/ShaoLY20,DBLP:conf/cvpr/JiaZSC20}.

\subsection{Implementation Details}
Following \cite{Fang_2022_WACV,DBLP:conf/icb/GeorgeM19,fairness_face_pad}, we sampled evenly 25 frames per video across the duration of each video in O, C, I, and M datasets. Then, we used MTCNN \cite{DBLP:journals/spl/ZhangZLQ16} to detect face for each frame and resized each face to $256 \times 256 \times 3$ pixels. In each mini-batch, we sampled the training data to keep a bona fide-attack ratio of 1:1 \cite{DBLP:conf/ai4i/ShimizuAOMHK18}. Our feature extractor is ResNet-18 initialized by the pre-trained weight on ImageNet \cite{DBLP:conf/cvpr/DengDSLL009}. The initial learning of feature extractor and classifier were set to 0.001 and 0.01, respectively. The learning rates were halved at 30 and 45 epochs with a maximum epoch of 60. During the training process, we used the Stochastic Gradient Descent (SGD) optimizer with a momentum of 0.9 and weight decay of 5e-4. The loss weight $\lambda$ was empirically set to $2$ for balancing two losses. In the inference phase, a final PAD decision score of a video is a fused score (mean-rule fusion) of all frames by following \cite{Fang_2022_WACV,DBLP:conf/icb/GeorgeM19,DBLP:journals/tbbis/YuLSXZ21,DBLP:journals/pami/YuWQLLZ21}. The hyper-parameters of CGMixStyle followed the setting of MixStyle \cite{DBLP:conf/iclr/ZhouY0X21} for a fair comparison. The hyper-parameters of CI was investigated in Sec. \ref{ssec:ablation_studies}.

\subsection{Evaluation Metrics}
Following existing cross-domain face PAD methods \cite{DBLP:conf/icmcs/LiuCDLZX22,Fang_2022_WACV,DBLP:conf/cvpr/LiPWK18,DBLP:conf/cvpr/ShaoLLY19,DBLP:conf/aaai/ShaoLY20}, we report the Half Total Error Rate (HTER), which is the mean of Bona fide Presentation Classification Error Rate (BPCER) \cite{ISO301073} and Attack Presentation Classification Error Rate (APCER) \cite{ISO301073} and Area under the Receiver Operating Characteristic (ROC) Curve (AUC) value for comparison.


\vspace{-2mm}
\section{Results} 
\label{sec:results}

\subsection{Analytical Experiments}
\label{ssec:ablation_studies}

\begin{table*}[htb!]
\begin{center}
\resizebox{0.95\textwidth}{!}{
\begin{tabular}{l|cc|cc|cc|cc|cc}
\hline
\multirow{2}{*}{Method} & \multicolumn{2}{c|}{O\&C\&I → M} & \multicolumn{2}{c|}{O\&M\&I → C} & \multicolumn{2}{c|}{O\&C\&M → I} & \multicolumn{2}{c|}{I\&C\&M → O} & \multicolumn{2}{c}{Average} \\ 
 & HTER(\%) $\downarrow$ & AUC(\%) $\uparrow$ & HTER(\%) $\downarrow$ & AUC(\%) $\uparrow$  & HTER(\%) $\downarrow$ & AUC(\%) $\uparrow$ & HTER(\%) $\downarrow$ & AUC(\%) $\uparrow$  & HTER(\%) $\downarrow$ & AUC(\%) $\uparrow$  \\ \hline \hline
w/o MixStyle             & 15.48       & 91.50       & 14.67       & \textbf{93.17}       & 22.70       & 81.63       & 16.77       & 91.03       & 17.41        & 89.33        \\
cross-domain              & 11.67       & 94.93       & 14.00       & 92.20       & 21.65       & 83.45       & 13.61       & 93.71       & 15.23        & 91.07        \\
random                   & \textbf{10.71}       & 94.71       & 19.00       & 88.41       & 22.55       & 83.37       & 13.64       & 93.57       & 16.48        & 90.02        \\  \hline
CGMixStyle (ours)             & 10.95       & \textbf{95.34}       & \textbf{12.56}       & 93.15       & \textbf{20.95}       & \textbf{85.88}       & \textbf{12.10}       & \textbf{95.07}       & \textbf{14.14}        & \textbf{92.36} 
\\ \hline
\end{tabular}}
\vspace{-2mm}
\caption{Comparison results of our CGMixStyle and MixStyle. cross-domain and random shuffle operations are presented by MixStyle \cite{DBLP:conf/iclr/ZhouY0X21}, while our CGMixStyle leverages the class-guided shuffle. The CGMixStyle obtains the best average performance, demonstrating the importance of the class-guided operation.}
\label{tab:shuffle_operations}
\vspace{-5mm}
\end{center}
\end{table*}

\paragraph{Comparison of CGMixStyle and MixStyle:} Tab. \ref{tab:shuffle_operations} compares the results of MixStyle with random and corss-domain shuffling operation and our CGMixStyle with class-guided shuffle.
The results in Tab. \ref{tab:shuffle_operations} show that the random Mixstyle fails to improve the results in one case. For example, in the case O\&M\&I → C, the HTER value is increased from 14.67\% (without MixStyle) to 19.00\% (with random MixStyle). Moreover, CGMixStyle outperform the MixStyle using random and cross-domain shuffle operations. The results suggest that the class-guided operation helps to enrich the style of the training data without confusing bona fide and attack patterns.

\begin{table*}[ht]
\begin{center}
\resizebox{0.95\textwidth}{!}{
\begin{tabular}{l|cc|cc|cc|cc|cc}
\hline
\multirow{2}{*}{Method} & \multicolumn{2}{c|}{O\&C\&I → M} & \multicolumn{2}{c|}{O\&M\&I → C} & \multicolumn{2}{c|}{O\&C\&M → I} & \multicolumn{2}{c|}{I\&C\&M → O} & \multicolumn{2}{c}{Average} \\ 
 & HTER(\%) $\downarrow$ & AUC(\%) $\uparrow$ & HTER(\%) $\downarrow$ & AUC(\%) $\uparrow$  & HTER(\%) $\downarrow$ & AUC(\%) $\uparrow$ & HTER(\%) $\downarrow$ & AUC(\%) $\uparrow$  & HTER(\%) $\downarrow$ & AUC(\%) $\uparrow$  \\ \hline \hline
 Baseline (w/ CGMixStyle)       & {10.95} & 95.34 & {12.56} & 93.15 & {20.95} & 85.88 & {12.10} & 95.07 & {14.14} & 92.36 \\ \hline
+ Random Zero                  & {9.52}  & 94.00 & {11.89} & 94.23 & {18.30} & 87.90 & {12.80} & 94.15 & {13.13} & 92.57 \\ 
+ Random Replace               & {8.81}  & 96.12 & {14.56} & 93.12 & {18.15} & 86.05 & {12.12} & 93.89 & {13.41} & 92.30 \\ 
+ Random Shuffle               & {8.33}  & \textbf{97.12} & {12.67} & 93.25 & {19.55} & 89.50 & {12.39} & 94.69 & {13.24} & 93.64 \\ \hline
All      & \textbf{8.11}        & 96.43       & \textbf{11.78}       & \textbf{95.64}       & \textbf{16.50}       & \textbf{91.50}      & \textbf{9.87}     &  \textbf{95.13}    & \textbf{11.57}        & \textbf{94.68}       \\ \hline
\end{tabular}}
\vspace{-2mm}
\caption{Effect of different types of CIs. All refers to the random execution of one intervention selected from random zero/replace/shuffle in each training mini-batch. 'All' outperforms other single CIs, proving that increasing the diversity of interventions helps model optimization. }
\label{tab:CI-operations}
\vspace{-8mm}
\end{center}
\end{table*}

\vspace{-4mm}
\paragraph{Effect of different types of CIs:} We investigated the effect of using three CIs to produce counterfactual features. The results are presented in Tab. \ref{tab:CI-operations}, where 'All' refers to randomly select one from three interventions (random zero, random replace, random shuffle) in each training mini-batch. The results indicate that applying a combination of different interventions outperform solutions using single intervention. This is consistent with the theoretical basis in Sec. \ref{ssec:background} that increasing diversity of counterfactual interventions can enhance the PAD generalizability. 

\begin{figure}[thbp!]
\begin{center}
\includegraphics[width=0.65\linewidth]{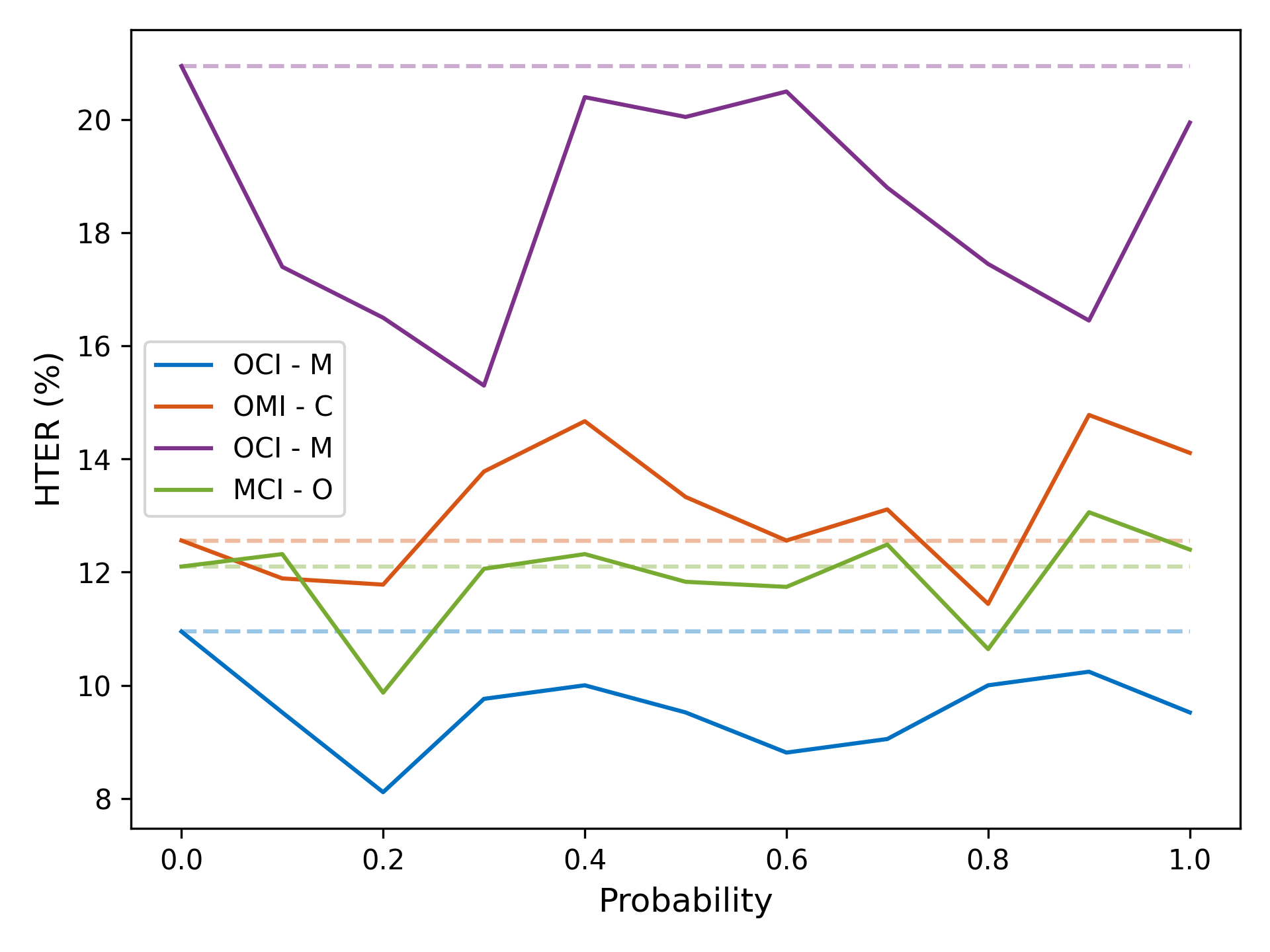}
\vspace{-2mm}
\caption{Effect of Hyperparameter in CIs. x-axis indicates the degree of counterfactual intervention in features and y-axis is the HTER (\%). Performing CI with any degree enhances the PAD performance in most cases, as solid lines underneath dashed lines of the same colour. The best overall performance is achieved mostly at $0.2$.}
\label{fig:hter_prob}
\end{center}
\vspace{-7mm}
\end{figure}

\vspace{-4mm}
\paragraph{Effect of Hyperparameter in CIs:} In addition to demonstrate the effectiveness of the different types of CIs, we conducted a set of experiments to analyze the sensitivity of hyperparameter of CIs. This hyperparameter control the feature destruction degree of CIs, i.e., percentage of random zeroing or replacing. 
For example, $0.2$ in x-axis in Fig. \ref{fig:hter_prob} indicates that $20\%$ values in the feature $Z$ (of dimension $512$) are manipulated by either setting it to zero or replacing it randomly with the value of the previous neighbour. $0.0$ indicates no intervention is performed, representing also as dashed lines. In most cases, performing CI using any hyperparameters achieved better PAD performance than no CI (i.e. solid lines underneath dashed lines of the same colour). Moreover, PAD performance achieves its best (lowest HTER values) mostly at $0.2$. 
In the following experiments, we therefore set this hyperparamters at $0.2$.

\vspace{-3mm}

\begin{table*}[htb!]
\begin{center}
\resizebox{0.95\textwidth}{!}{
\begin{tabular}{l|cc|cc|cc|cc|cc}
\hline
\multirow{2}{*}{Method} & \multicolumn{2}{c|}{O\&C\&I → M} & \multicolumn{2}{c|}{O\&M\&I → C} & \multicolumn{2}{c|}{O\&C\&M → I} & \multicolumn{2}{c|}{I\&C\&M → O} & \multicolumn{2}{c}{Average} \\ 
 & HTER(\%) $\downarrow$ & AUC(\%) $\uparrow$ & HTER(\%) $\downarrow$ & AUC(\%) $\uparrow$  & HTER(\%) $\downarrow$ & AUC(\%) $\uparrow$ & HTER(\%) $\downarrow$ & AUC(\%) $\uparrow$  & HTER(\%) $\downarrow$ & AUC(\%) $\uparrow$  \\ \hline \hline
Baseline                & {15.48} & 91.50 & {14.67} & 93.17 & {22.70} & 81.63 & {16.77} & 91.03 & {17.41} & 89.33 \\ \hline
+ CGMixStyle             & {10.95} & 95.34 & {12.56} & 93.15 & {20.95} & 85.88 & {12.10} & 95.07 & {14.14} & 92.36 \\ 
+ CI        & {9.29}  & 94.35 & {12.44} & 93.89 & {18.50}  & 85.61 & {13.74} & 90.01 & {13.49} & 90.97 \\ \hline
CF-PAD (all)       & \textbf{8.11}        & \textbf{96.43}       & \textbf{11.78}       & \textbf{95.64}       & \textbf{}16.50       & \textbf{91.50}     & \textbf{9.87}     &  \textbf{95.13}    & \textbf{11.57}        & \textbf{94.68}       \\ \hline
\end{tabular}}
\caption{Ablation studies on two components, CGMixStyle and CI, in our CF-PAD solution. The bold number indicates the best performance for each evaluation protocol, demonstrating the importance of both components in CF-PAD.}
\vspace{-6mm}
\label{tab:ablation-results}
\end{center}
\end{table*}

\vspace{-4mm}
\paragraph{Effect of CGMixStyle and CIs in CF-PAD:} After a detailed analysis of CGMixStyle and CI, we discuss the influence of these two modules in the proposed CF-PAD approach, as the results presented in Tab. \ref{tab:ablation-results}. We have the following observations: 1) both module enhances the face PAD generalizability, respectively. 2) CF-PAD, combining the CGMixStyle and CI, achieves the best performance. The results point out that two modules are mutually complementary and strengthening, consisting with the theoretical rationale in Sec. \ref{ssec:background}.

\vspace{-2mm}
\subsection{Comparison with State-of-the-Art methods} 
The following presents the achieved performances on the considered evaluations in comparison to the SOTA that reported on each setup to capture the widest range of comparisons and the different levels of training data availability. 
\begin{table*}[]
\begin{center}
\resizebox{0.95\textwidth}{!}{
\begin{tabular}{l|cc|cc|cc|cc|cc}
\hline
\multirow{2}{*}{Method} & \multicolumn{2}{c|}{O\&C\&I → M} & \multicolumn{2}{c|}{O\&M\&I → C} & \multicolumn{2}{c|}{O\&C\&M → I} & \multicolumn{2}{c|}{I\&C\&M → O} & \multicolumn{2}{c}{Average} \\ 
 & HTER(\%) $\downarrow$ & AUC(\%) $\uparrow$ & HTER(\%) $\downarrow$ & AUC(\%) $\uparrow$  & HTER(\%) $\downarrow$ & AUC(\%) $\uparrow$ & HTER(\%) $\downarrow$ & AUC(\%) $\uparrow$  & HTER(\%) $\downarrow$ & AUC(\%) $\uparrow$  \\ \hline \hline
Binary CNN  \cite{DBLP:conf/eccv/XuLNX14}               & 29.25       & 82.87       & 34.88       & 71.94       & 34.47       & 65.88       & 29.61       & 77.54       & 32.05        & 74.56        \\
Auxiliary  \cite{DBLP:conf/cvpr/LiuJ018} & 22.72       & 85.88       & 33.52       & 73.15       & 29.14       & 71.69       & 30.17       & 77.61       & 28.89        & 77.08        \\
ResNet50-PS \cite{DBLP:journals/tbbis/YuLSXZ21} & 14.32       & 94.51       & 18.23       & 89.75       & 18.86       & 89.63       & 21.44       & 87.56       & 18.21        & 90.36        \\
NAS-FAS \cite{DBLP:journals/pami/YuWQLLZ21}   & 19.53       & 88.63       & 16.54       & 90.18       & 14.51       & 93.84       & 13.8        & 93.43       & 16.10        & 91.52        \\
LMFD \cite{Fang_2022_WACV}  & 10.48       & 94.55       & 12.50       & 94.17       & 18.49       & 84.72       & 12.41       & 94.95       & 13.47        & 92.10        \\
ViTransPAD \cite{DBLP:conf/icip/MingYAVLB22} & 8.39        & -         & 21.27       & -         & 16.83       & -         & 15.63       & -          & 15.53        & -            \\
PatchNet \cite{DBLP:conf/cvpr/WangLYL22}  & 7.10        & 98.64       & 11.33       & 94.58       & 14.60       & 92.51       & 11.82       & 95.07       & 11.21        & 95.20        \\ \hline
MADDG \cite{DBLP:conf/cvpr/ShaoLLY19}        & 17.69       & 88.06       & 24.50       & 84.51       & 22.19       & 84.99       & 27.89       & 80.02       & 23.07        & 84.40        \\
RFM \cite{DBLP:conf/aaai/ShaoLY20}        & 17.30       & 90.48       & 13.89       & 93.98       & 20.27       & 88.16       & 16.45       & 91.16       & 16.98        & 90.95        \\
SSDG-R \cite{DBLP:conf/cvpr/JiaZSC20} & 7.38        & 97.17       & 10.44       & 95.94       & 11.71       & 96.59       & 15.61       & 91.54       & 11.29        & 95.31        \\
D$^2$AM \cite{DBLP:conf/aaai/ChenYSDTLHJ21} & 12.70 & 95.66 &  20.98 & 85.58 & 15.43 & 91.22 & 15.27 & 90.87 & 16.10 & 90.83 \\
ViT  \cite{DBLP:conf/eccv/HuangSLCXYAY22} & \textbf{4.75}        & \textbf{98.59}       & 15.70       & 92.76       & 17.68       & 86.66       & 16.46       & 90.37       & 13.65        & 92.10        \\
TransFAS \cite{DBLP:journals/tbbis/WangWDG22}    & 7.08        & 96.69       & 9.81        & 96.13       & 10.12       & 95.53       & 15.52       & 91.10       & 10.63        & 94.86        \\
DADN-CDS \cite{DBLP:journals/tcsv/YanZH22}  & 5.24        & 98.06       & \textbf{6.84}        & \textbf{97.95}       & 10.64       & 95.14       & 13.77       & 93.09       & \textbf{9.12}         & \textbf{96.06}        \\
CIFAS \cite{DBLP:conf/icmcs/LiuCDLZX22}        & 5.95        & 96.32       & 10.66       & 95.30       & \textbf{8.50}        & \textbf{97.24}       & 13.17       & 93.44       & 9.57         & 95.58        \\ \hline
Baseline      & 15.48       & 91.50       & 14.67       & 93.17     & 22.70       & 81.63       & 16.77       & 91.03       & 17.41        & 89.33        \\
CF-PAD (ours)                     & 8.11        & 96.43       & 11.78       & 95.64       & 16.50       & 91.50      & \textbf{9.87}        & \textbf{95.13}       & 11.57        & 94.68       \\ \hline
\end{tabular}}
\vspace{-2mm}
\caption{The results of the four multi-source cross-dataset evaluations comparing our CF-PAD and SOTA PAD solutions. Methods in the first block are conventional PAD solution (i.e., no domain adaptation), while methods in the second block are explicitly designed to target the domain shift problem. The bold number indicates the best performance under each protocol. Our CF-PAD demonstrates a very competitive PAD performance.}
\label{tab:sota_multiple_cross_db}
\end{center}
\vspace{-6mm}
\end{table*}

\begin{table}[]
\tiny
\begin{center}
\resizebox{0.35\textwidth}{!}{
\begin{tabular}{l|cc}
\hline
Method           & HTER(\%) $\downarrow$ & AUC(\%)  \\ \hline
GRL Layer \cite{DBLP:conf/icml/GaninL15}       & 29.1 & 76.4 \\ 
ADDA \cite{DBLP:conf/cvpr/TzengHSD17} & 33.7 & 70.3 \\ 
DA-FAS \cite{DBLP:conf/cvpr/SahaXKGCPG20}       & 27.1 & 79.2 \\ 
UCDA-FAS \cite{DBLP:conf/fgr/PanwarSSPG21}          & 26.1 & 80.0 \\ 
CIFAS \cite{DBLP:conf/icmcs/LiuCDLZX22}            & 24.6 & 83.2 \\ \hline
Baseline & 27.1 & 80.3 \\ 
CF-PAD (ours)    & \textbf{23.5} & \textbf{84.2} \\  \hline
\end{tabular}}
\vspace{-2mm}
\caption{Comparison results of models trained on the O\&C\&M and tested on the large-scale CA. }
\label{tab:ocm_ca}
\vspace{-5mm}
\end{center}
\end{table}

\begin{table}[]
\small
\begin{center}
\resizebox{0.47\textwidth}{!}{
\begin{tabular}{l|cc|cc}
\hline
\multirow{2}{*}{Method} & \multicolumn{2}{c|}{M\&I → C} & \multicolumn{2}{c}{M\&I → O} \\
 & HTER(\%) $\downarrow$ & AUC(\%) & HTER(\%) $\downarrow$ & AUC(\%)      \\
\hline
MS-LBP \cite{DBLP:conf/icb/MaattaHP11}               & 51.16        & 52.09       & 43.63        & 58.07       \\
IDA \cite{DBLP:journals/tifs/WenHJ15}                    & 45.16        & 58.80       & 54.52        & 42.17       \\
MADDG \cite{DBLP:conf/cvpr/ShaoLLY19}                & 41.02        & 64.33       & 39.35        & 65.10       \\
RFM \cite{DBLP:conf/aaai/ShaoLY20}                      & 36.34        & 67.52       & 29.12        & 72.61       \\
SSDG-R  \cite{DBLP:conf/cvpr/JiaZSC20}   & 31.89        & 71.29       & 36.01        & 66.88       \\
DR-MD-Net \cite{DBLP:conf/cvpr/Wang0SC20}  & 31.67        & 75.23       & 34.02        & 72.65       \\
D$^2$AM \cite{DBLP:conf/aaai/ChenYSDTLHJ21} & 32.65 & 72.04 & 27.70 & 75.36  \\
CIFAS \cite{DBLP:conf/icmcs/LiuCDLZX22}  & 22.67        & 83.39       & 24.63        & 81.48       \\ \hline
Baseline & 25.78 & 80.56 & 23.24 & 84.66 \\
CF-PAD (ours)                    & \textbf{22.11}        & \textbf{85.06}       & \textbf{19.71}        & \textbf{89.01}  \\ \hline
\end{tabular}}
\vspace{-2mm}
\caption{Limited-source cross-dataset results in terms of HTER (\%) and AUC (\%).}
\label{tab:mi_c_o}
\vspace{-9mm}
\end{center}
\end{table}

\vspace{-5mm}
\paragraph{Multi-source cross-dataset evaluation:} Tab. \ref{tab:sota_multiple_cross_db} shows results of the following four evaluation protocols: O\&C\&I → M, O\&M\&I → C, O\&C\&M → I, I\&C\&M → O, by comparing with several SOTA face PAD solutions. 
From Tab. \ref{tab:sota_multiple_cross_db}, we observe that: 1) The proposed CF-PAD approach outperforms baseline method (without CGMixStyle and CI) by a drop of 5.84 percentage points on HTER and an enhancement of 5.35 percentage points on AUC. 2) Our proposed CF-PAD method obtains competitive and even better PAD performance than SOTA approaches. Under the challenging case of I\&C\&M → O, our CF-PAD approach outperforms the SOTA methods with an HTER value of 9.87\% and an AUC value of 95.13\%. 

Furthermore, Tab. \ref{tab:ocm_ca} shows the results of most challenging case O\&C\&M → CA, as CA poses a high diversity in terms of the attack types, identity, and capture environments. Our CF-PAD solution achieves the best performance in comparison to the SOTA methods. Both results suggest that our CF-PAD method can identify the causal and generalized discriminative features for cross-dataset scenarios.

\vspace{-4mm}
\paragraph{Limited-source cross-dataset evaluation:} Tab. \ref{tab:mi_c_o} shows the results of models trained on the limited two training datasets. Our CF-PAD approach outperforms the SOTA face PAD methods, further proving the generalizability of CGMixStyle and CI modules.

\begin{figure}[thbp!]
\vspace{-3mm}
\begin{center}
\includegraphics[width=0.68\linewidth]{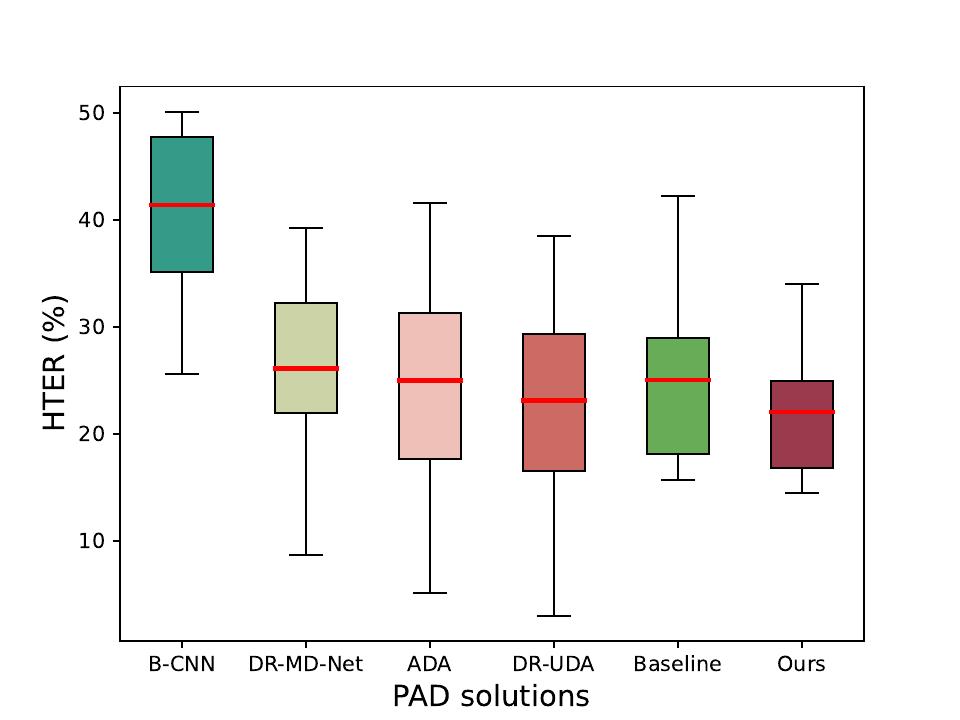}
\vspace{-2mm}
\caption{The Box and Whisker plot of the performance variation under single-source cross-dataset scenarios reported in Tab. \ref{tab:singe-db-results}. Each box refers to a model evaluated on a total of 12 protocols (B-CNN is binary CNN method). The red line within the box is the mean value of 12 testing cases, the box range represents the standard deviation, and the top straight line extended of the box is the worst-case. Note the lowest average HTER value of our CF-PAD, along with its low deviation and the optimal worst-case.}
\label{fig:single_db_box_plot}
\vspace{-5mm}
\end{center}
\end{figure}

\vspace{-4mm}
\paragraph{Single-source cross-dataset evaluation:} To further verify the effectiveness of the proposed method, we evaluate the CF-PAD under the highly challenging scenario where only one dataset is available for training. Tab. \ref{tab:singe-db-results} and Fig. \ref{fig:single_db_box_plot} presents the results of our method compared to the SOTA method. Our method improves the PAD generalizability in most cases and achieves the lowest average error rates. In addition, CF-PAD improves the worst case, where the highest HTER value (worst) among 12 cases is 34.00\% obtained by CF-PAD and 38.50\% by DR-UDA \cite{DBLP:journals/tifs/WangHSC21}. These results implies that enriching the diversity of feature-level training data and the diversity of CIs leads to a better causal and representative feature learning as rationalized by Eq. \ref{eq:erm_scm}. Fig. \ref{fig:single_db_box_plot} illustrates the mean and standard deviation of the HTER values of the results showed in Tab. \ref{tab:singe-db-results}. The plot stresses that CF-PAD performs more consistently (the lowest average error rates, the lowest deviation, and the lowest errors of the worst-case) than other PAD solutions in the challenging scenarios.

\begin{table*}[]
\begin{center}
\resizebox{0.98\textwidth}{!}{
\begin{tabular}{l|ccc|ccc|ccc|ccc||cc}
\hline
Method               & C → I   & C → M   & C → O  & I → C   & I → M   & I → O  & M → C   & M → I   & M → O   & O → I   & O → M   & O→  C   & Average  & Worst \\ \hline
Binary CNN \cite{DBLP:journals/corr/YangLL14}   & 45.80 & 25.60  & 36.40 & 44.40 & 48.60 & 45.40 & 50.10 & 49.90  & 31.40 & 47.40 & 30.20 & 41.20  & 41.37 ± 8.42 & 50.01 \\ 
ADA \cite{DBLP:conf/icb/WangHSC19}     & 17.50 & 9.30  & 29.10 & 41.60 & 30.50 & 39.60 & 17.70 & 5.10 & 31.20  & 26.80 & 31.50 & 19.80   & 24.98 ± 11.28 & 41.60 \\ 
DR-MD-Net \cite{DBLP:conf/cvpr/Wang0SC20} & 26.10 & 20.20 & 24.70 & 39.20 & 23.20 & 33.60  & 34.30 & 8.70 & 31.70  & 27.60 & 22.00 & 21.80   & 26.09 ± 7.70 & 39.20  \\ 
DR-UDA \cite{DBLP:journals/tifs/WangHSC21}     & \textbf{15.60} & \textbf{9.00} & 28.70 & 34.20 & 29.00 & 38.50  & 16.80 & \textbf{3.00} & 30.20 & 25.40 & 27.40 & \textbf{19.50}   & 23.11 ± 10.50 & {38.50} \\ \hline
Baseline             & 38.85 & 18.10 & \textbf{17.94} & 42.22 & \textbf{18.81} & \textbf{28.42} & 27.11 & 16.30  & 30.49 & 23.10 & 15.71 & 23.11 & 25.01 ± \textit{8.74} & 42.22\\ 
CF-PAD (ours)                 & 24.80 & 17.14 & 19.43 & \textbf{34.00} & 24.76 & 31.70  & \textbf{14.44} & 15.90 & \textbf{25.34} & \textbf{21.50} & \textbf{15.00} & 20.33  & \textbf{22.03} ± \textbf{6.33} & \textbf{34.00} \\ \hline
\end{tabular}}
\vspace{-2mm}
\caption{Single-source cross-dataset results in terms of HTER (\%). The bold number indicates the lowest error rates, indicating a better generalizability of our CF-PAD.}
\label{tab:singe-db-results}
\vspace{-7mm}
\end{center}
\end{table*}

\vspace{-3mm}
\begin{table}[]
\tiny
\begin{center}
\resizebox{0.47\textwidth}{!}{
\begin{tabular}{l|ccc}
\hline
Methods     & Param. (M) & FLOPs (G) & HTER (\%) \\ \hline
Auxiliary \cite{DBLP:conf/cvpr/LiuJ018}  & \textbf{2.20}           & 47.41     & 28.89   \\ 
ResNet50-PS \cite{DBLP:journals/tbbis/YuLSXZ21}  & 23.54          & 24.66     & 18.21        \\ 
NAS-FAS \cite{DBLP:journals/pami/YuWQLLZ21}  & {2.94}  & 52.67     & 16.10        \\ 
LMFD  \cite{Fang_2022_WACV}       & 101.72         & 36.11     & 13.47        \\ 
ViTransPAD \cite{DBLP:conf/icip/MingYAVLB22}  & 66.00          & 7.88      & 15.22        \\ 
MADDG \cite{DBLP:conf/cvpr/ShaoLLY19}       & 17.81          & 189.61    & 23.07        \\ 
RFM  \cite{DBLP:conf/aaai/ShaoLY20}       & 3.87           & 95.79     & 16.89        \\ 
SSDG-R \cite{DBLP:conf/cvpr/JiaZSC20}     & 12.23          & 14.26     & 11.29        \\ 
ViT \cite{DBLP:conf/eccv/HuangSLCXYAY22}      & 86.39          & 33.69     & 13.65        \\ 
DADN-CDS \cite{DBLP:journals/tcsv/YanZH22}   & 23.61          & 56.00     & \textbf{9.12}         \\ \hline
CF-PAD (ours)         & 11.18          & \textbf{1.82}      & 11.57        \\ \hline
\end{tabular}}
\vspace{-2mm}
\caption{Computational complexity vs. PAD Performance. HTER (\%) refers to the average HTER values reported in Tab. \ref{tab:sota_multiple_cross_db}. The proposed CF-PAD achieves the lowest complexity (FLOPs) and a very competitive PAD performance.}
\label{tab:computation_complexity}
\vspace{-9mm}
\end{center}
\end{table}

\subsection{Performance vs. Computational Complexity}
\label{ssec:complexity}
To enable the efficiency comparison, we account the trainable parameters of models and the floating point operations (FLOPs) of our method and the PAD works listed in Tab. \ref{tab:sota_multiple_cross_db} when it is feasible.
Tab. \ref{tab:computation_complexity} and Fig. \ref{fig:computation_complexity} demonstrate the efficiency of the propose CF-PAD method by reporting the performance as the HTER, and computational complexity in terms of the number of trainable parameters and the FLOPs. Five PAD methods in Tab. \ref{tab:sota_multiple_cross_db}  were ignored for efficiency comparison due to no open source implementation and no detailed model architecture available for accounting. Fig. \ref{fig:computation_complexity} presents the number of FLOPs (x-axis) vs. the PAD performance (y-axis) vs. the trainable parameters (radius of the circle) of our method and the SOTA PADs. The ideal solution will tend to be placed on the bottom left corner (low error rates and low complexity) as well as small circle (less trainable parameters). The proposed CF-PAD method achieves the lowest FLOPs and very competitive PAD performance, indicating its relative high efficiency. The detailed values presented in Tab. \ref{tab:computation_complexity} supports this conclusion.
\begin{figure}[th!]
\begin{center}
\begin{subfigure}[b]{0.48\linewidth}
     \centering
     \includegraphics[width=\linewidth]{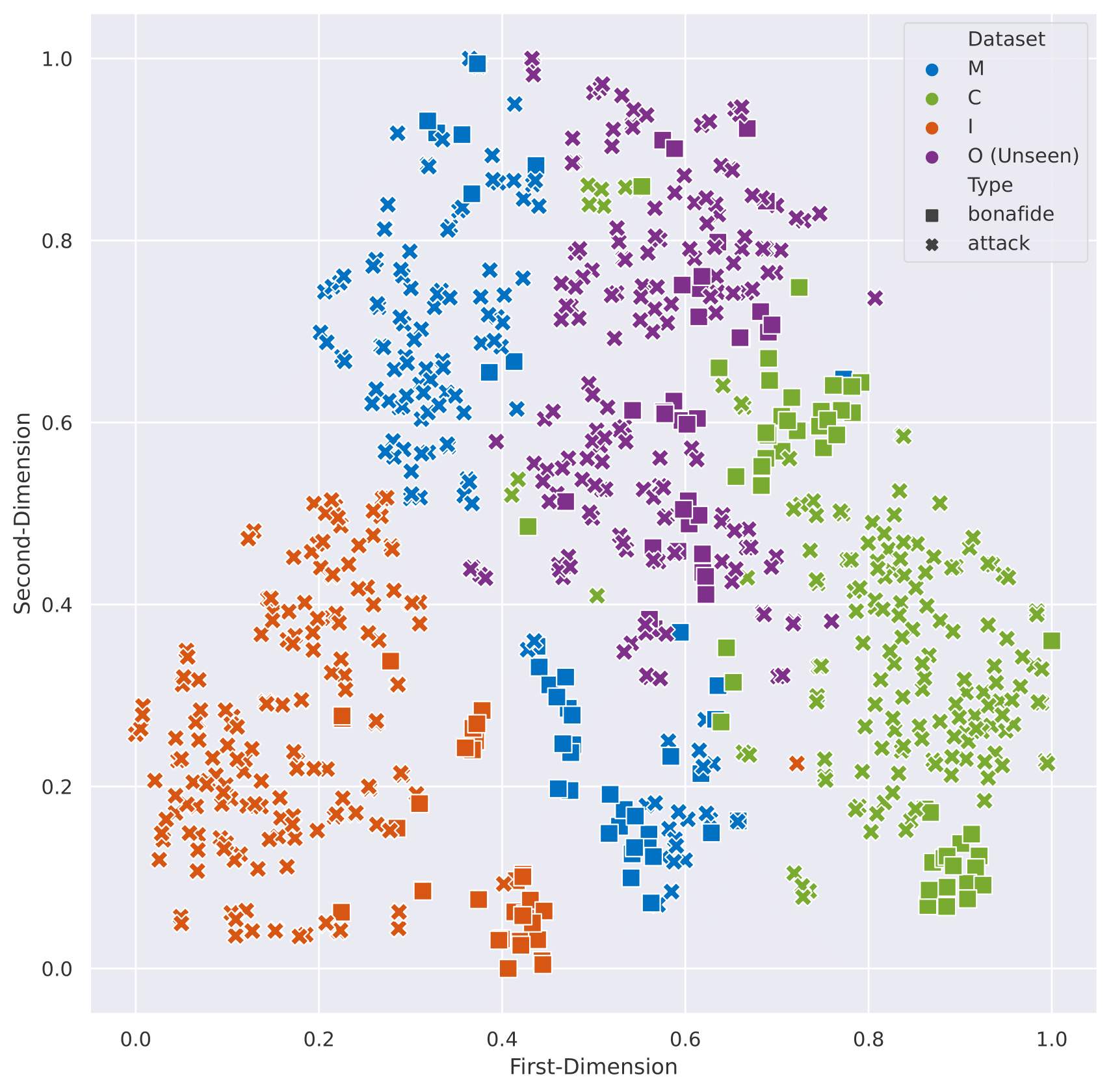}
     \caption{Baseline}
\end{subfigure}
\begin{subfigure}[b]{0.48\linewidth}
     \centering
     \includegraphics[width=\linewidth]{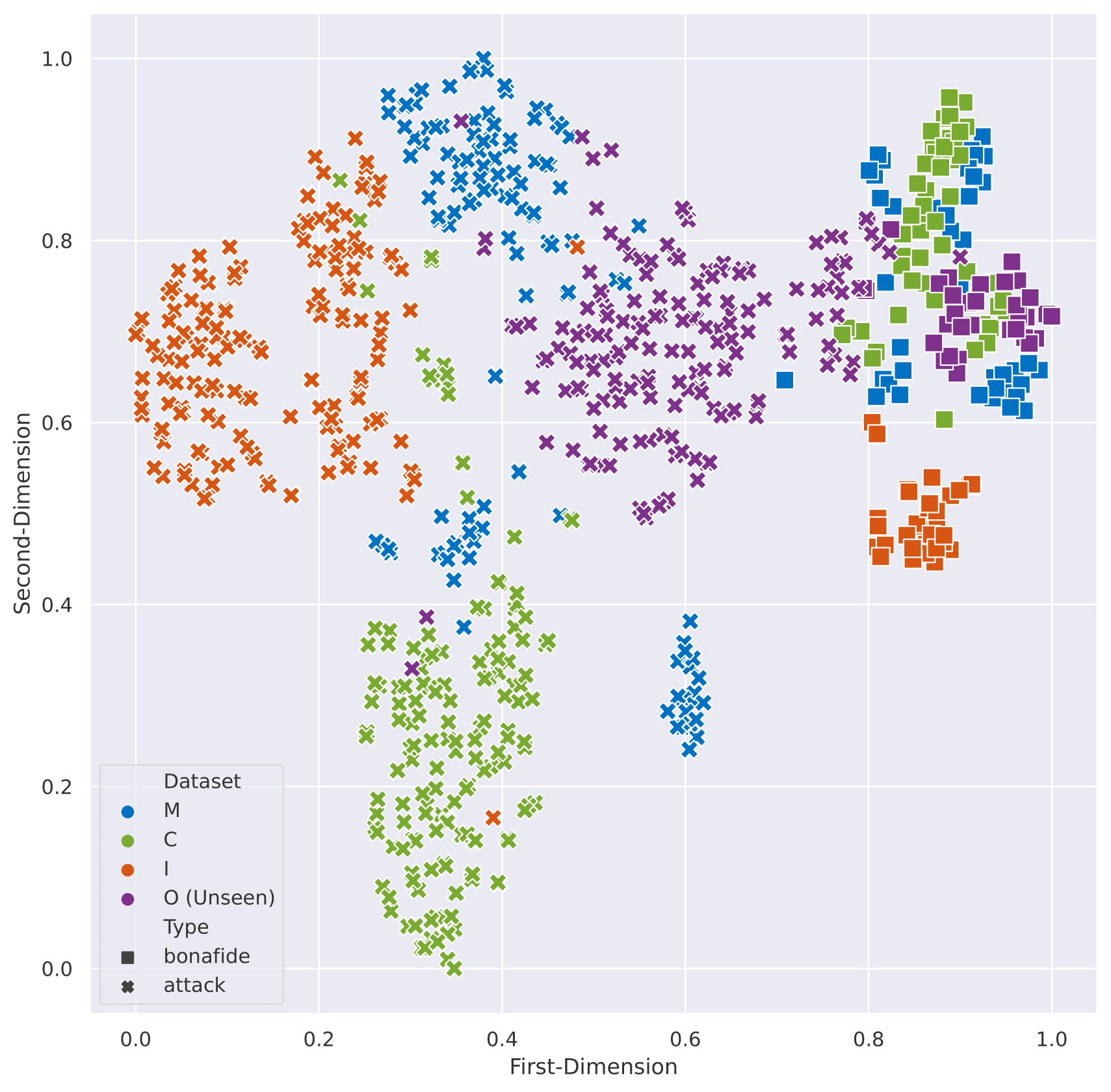}
     \caption{CF-PAD (ours)}
\end{subfigure}
\vspace{-2mm}
\caption{Feature visualization of the case I\&C\&M → O by baseline (a) and our CF-PAD solution (b), respectively. Different colors indicates the different dataset, where O (purple) is unseen testing data, and different shape represents the different classes, where square is bona fide and cross is attack. It is clear that CF-PAD offers a more discriminative feature space.}
\label{fig:tsnes}
\vspace{-9mm}
\end{center}
\end{figure}

\vspace{-4mm}
\subsection{Visualization and Analysis}
We visualized the feature distribution learned by the baseline and our CF-PAD model under the protocol I\&C\&M → O in Fig. \ref{fig:tsnes}. To make a clear observation and avoid possible overlapping region, we randomly select 300 samples from each dataset and illustrate their distribution by using t-SNE \cite{JMLR:v9:vandermaaten08a}. Each color in the plots represent the different datasets (purple is unseen testing dataset OULU-NPU) and each shape refers to different classes (square and cross represent bona fide and attack samples, respectively).
Comparing Fig. \ref{fig:tsnes} (a) and (b), we observe that samples from the same dataset (same color) are more clustered by the baseline model and sample from the same classes (same shape) are more clustered by our CF-PAD, especially for the unseen O data. This observation proves that our CF-PAD learns a causal and generalized discriminative feature space for face PAD instead of focusing on features from different domains. Moreover, this observation implies that the proposed CF-PAD without accessing the target testing data is effective and reliable for complex practical scenarios.

\vspace{-3mm}
\section{Conclusion}
This paper models the face PAD task from a causal view, aiming to enhance the generalizability of PAD under cross-dataset scenarios.
The main idea is to approximate the model optimization from two aspects, 1) enriching the feature-level training data diversity via CGMixStyle and 2) performing a set of CIs on high-level features, to learn a causal and generalized feature representation. In addition, CGMixStyle and CI components introduce no extra trainable parameters, negligible computation complexity during training, and no computation resources in the inference phase. 
A wide set of experiments and comparisons reveal the effectiveness and efficiency of our proposed method.

\vspace{-5mm}
\paragraph{Acknowledgment:}
This research work has been funded by the German Federal Ministry of Education and Research and the Hessen State Ministry for Higher Education, Research and the Arts within their joint support of the National Research Center for Applied Cybersecurity ATHENE.

{\small
\bibliographystyle{ieee_fullname}
\bibliography{egbib}

\begin{thebibliography}{10}\itemsep=-1pt

\bibitem{oulu_npu}
Zinelabdine Boulkenafet, Jukka Komulainen, Lei Li, Xiaoyi Feng, and Abdenour
  Hadid.
\newblock {OULU-NPU:} {A} mobile face presentation attack database with
  real-world variations.
\newblock In {\em {FG}}, pages 612--618. {IEEE} Computer Society, 2017.

\bibitem{DBLP:conf/cvpr/BoutrosDKK22}
Fadi Boutros, Naser Damer, Florian Kirchbuchner, and Arjan Kuijper.
\newblock Elasticface: Elastic margin loss for deep face recognition.
\newblock In {\em {CVPR} Workshops}, pages 1577--1586. {IEEE}, 2022.

\bibitem{BOUTROS2023104688}
Fadi Boutros, Vitomir Struc, Julian Fierrez, and Naser Damer.
\newblock Synthetic data for face recognition: Current state and future
  prospects.
\newblock {\em Image and Vision Computing}, 135:104688, 2023.

\bibitem{DBLP:conf/iccv/0002LL021}
Guangyi Chen, Junlong Li, Jiwen Lu, and Jie Zhou.
\newblock Human trajectory prediction via counterfactual analysis.
\newblock In {\em {ICCV}}, pages 9804--9813. {IEEE}, 2021.

\bibitem{DBLP:conf/aaai/ChenYSDTLHJ21}
Zhihong Chen, Taiping Yao, Kekai Sheng, Shouhong Ding, Ying Tai, Jilin Li,
  Feiyue Huang, and Xinyu Jin.
\newblock Generalizable representation learning for mixture domain face
  anti-spoofing.
\newblock In {\em {AAAI}}, pages 1132--1139. {AAAI} Press, 2021.

\bibitem{replay_attack}
Ivana Chingovska, Andr{\'{e}} Anjos, and S{\'{e}}bastien Marcel.
\newblock On the effectiveness of local binary patterns in face anti-spoofing.
\newblock In {\em {BIOSIG}}, volume {P-196} of {\em {LNI}}, pages 1--7. {GI},
  2012.

\bibitem{DBLP:conf/bmvc/DamerD16}
Naser Damer and Kristiyan Dimitrov.
\newblock Practical view on face presentation attack detection.
\newblock In {\em Proceedings of the British Machine Vision Conference 2016,
  {BMVC} 2016, York, UK, September 19-22, 2016}. {BMVA} Press, 2016.

\bibitem{DBLP:conf/cvpr/DengDSLL009}
Jia Deng, Wei Dong, Richard Socher, Li{-}Jia Li, Kai Li, and Fei{-}Fei Li.
\newblock Imagenet: {A} large-scale hierarchical image database.
\newblock In {\em {CVPR}}, pages 248--255. {IEEE} Computer Society, 2009.

\bibitem{DBLP:conf/cvpr/DengGXZ19}
Jiankang Deng, Jia Guo, Niannan Xue, and Stefanos Zafeiriou.
\newblock Arcface: Additive angular margin loss for deep face recognition.
\newblock In {\em {IEEE} Conference on Computer Vision and Pattern Recognition,
  {CVPR} 2019, Long Beach, CA, USA, June 16-20, 2019}, pages 4690--4699.
  Computer Vision Foundation / {IEEE}, 2019.

\bibitem{DBLP:conf/icb/FangAKD22}
Meiling Fang, Hamza Ali, Arjan Kuijper, and Naser Damer.
\newblock Patchswap: Boosting the generalizability of face presentation attack
  detection by identity-aware patch swapping.
\newblock In {\em {IJCB}}, pages 1--10. {IEEE}, 2022.

\bibitem{DBLP:conf/fgr/FangBKD21}
Meiling Fang, Fadi Boutros, Arjan Kuijper, and Naser Damer.
\newblock Partial attack supervision and regional weighted inference for masked
  face presentation attack detection.
\newblock In {\em 16th {IEEE} FG, Jodhpur, India, December 15-18, 2021}, pages
  1--8. {IEEE}, 2021.

\bibitem{Fang_2022_WACV}
Meiling Fang, Naser Damer, Florian Kirchbuchner, and Arjan Kuijper.
\newblock Learnable multi-level frequency decomposition and hierarchical
  attention mechanism for generalized face presentation attack detection.
\newblock In {\em {IEEE/CVF} WACV, Waikoloa, HI, USA, January 3-8, 2022}, pages
  1131--1140. {IEEE}, 2022.

\bibitem{DBLP:journals/pr/FangDKK22}
Meiling Fang, Naser Damer, Florian Kirchbuchner, and Arjan Kuijper.
\newblock Real masks and spoof faces: On the masked face presentation attack
  detection.
\newblock {\em Pattern Recognit.}, 123:108398, 2022.

\bibitem{DBLP:conf/cvpr/FangHD23}
Meiling Fang, Marco Huber, and Naser Damer.
\newblock Synthaspoof: Developing face presentation attack detection based on
  privacy-friendly synthetic data.
\newblock In {\em {CVPR} Workshops}, pages 1061--1070. {IEEE}, 2023.

\bibitem{fairness_face_pad}
Meiling Fang, Wufei Yang, Arjan Kuijper, Vitomir Struc, and Naser Damer.
\newblock Fairness in face presentation attack detection.
\newblock {\em CoRR}, abs/2209.09035, 2022.

\bibitem{DBLP:journals/ai/FentonNC20}
Norman~E. Fenton, Martin Neil, and Anthony~C. Constantinou.
\newblock The book of why: The new science of cause and effect, judea pearl,
  dana mackenzie. basic books {(2018)}.
\newblock {\em Artif. Intell.}, 284:103286, 2020.

\bibitem{DBLP:conf/icml/GaninL15}
Yaroslav Ganin and Victor~S. Lempitsky.
\newblock Unsupervised domain adaptation by backpropagation.
\newblock In {\em {ICML}}, volume~37 of {\em {JMLR} Workshop and Conference
  Proceedings}, pages 1180--1189. JMLR.org, 2015.

\bibitem{DBLP:conf/icb/GeorgeM19}
Anjith George and S{\'{e}}bastien Marcel.
\newblock Deep pixel-wise binary supervision for face presentation attack
  detection.
\newblock In {\em {ICB}}, pages 1--8. {IEEE}, 2019.

\bibitem{hagmayer2007causal}
York Hagmayer, Steven~A Sloman, David~A Lagnado, and Michael~R Waldmann.
\newblock Causal reasoning through intervention.
\newblock {\em Causal learning: Psychology, philosophy, and computation}, pages
  86--100, 2007.

\bibitem{DBLP:conf/eccv/HuangSLCXYAY22}
Hsin{-}Ping Huang, Deqing Sun, Yaojie Liu, Wen{-}Sheng Chu, Taihong Xiao,
  Jinwei Yuan, Hartwig Adam, and Ming{-}Hsuan Yang.
\newblock Adaptive transformers for robust few-shot cross-domain face
  anti-spoofing.
\newblock In {\em {ECCV} {(13)}}, volume 13673 of {\em Lecture Notes in
  Computer Science}, pages 37--54. Springer, 2022.

\bibitem{ISO301073}
{International Organization for Standardization}.
\newblock {ISO/IEC DIS 30107-3:2016: Information Technology – Biometric
  presentation attack detection – P. 3: Testing and reporting}, 2017.

\bibitem{DBLP:conf/cvpr/JiaZSC20}
Yunpei Jia, Jie Zhang, Shiguang Shan, and Xilin Chen.
\newblock Single-side domain generalization for face anti-spoofing.
\newblock In {\em {CVPR}}, pages 8481--8490. Computer Vision Foundation /
  {IEEE}, 2020.

\bibitem{DBLP:journals/tifs/LiLCWHK18}
Haoliang Li, Wen Li, Hong Cao, Shiqi Wang, Feiyue Huang, and Alex~C. Kot.
\newblock Unsupervised domain adaptation for face anti-spoofing.
\newblock {\em {IEEE} Trans. Inf. Forensics Secur.}, 13(7):1794--1809, 2018.

\bibitem{DBLP:conf/cvpr/LiPWK18}
Haoliang Li, Sinno~Jialin Pan, Shiqi Wang, and Alex~C. Kot.
\newblock Domain generalization with adversarial feature learning.
\newblock In {\em {CVPR}}, pages 5400--5409. {IEEE} Computer Society, 2018.

\bibitem{DBLP:conf/icmcs/LiuCDLZX22}
Yuchen Liu, Yabo Chen, Wenrui Dai, Chenglin Li, Junni Zou, and Hongkai Xiong.
\newblock Causal intervention for generalizable face anti-spoofing.
\newblock In {\em {ICME}}, pages 1--6. {IEEE}, 2022.

\bibitem{DBLP:conf/cvpr/LiuJ018}
Yaojie Liu, Amin Jourabloo, and Xiaoming Liu.
\newblock Learning deep models for face anti-spoofing: Binary or auxiliary
  supervision.
\newblock In {\em {CVPR}}, pages 389--398. {IEEE} Computer Society, 2018.

\bibitem{DBLP:conf/icb/MaattaHP11}
Jukka M{\"{a}}{\"{a}}tt{\"{a}}, Abdenour Hadid, and Matti Pietik{\"{a}}inen.
\newblock Face spoofing detection from single images using micro-texture
  analysis.
\newblock In {\em {IJCB}}, pages 1--7. {IEEE} Computer Society, 2011.

\bibitem{DBLP:conf/icip/MingYAVLB22}
Zuheng Ming, Zitong Yu, Musab Al{-}Ghadi, Muriel Visani, Muhammad~Muzzamil
  Luqman, and Jean{-}Christophe Burie.
\newblock Vitranspad: Video transformer using convolution and self-attention
  for face presentation attack detection.
\newblock In {\em {ICIP}}, pages 4248--4252. {IEEE}, 2022.

\bibitem{DBLP:conf/fgr/PanwarSSPG21}
Ankush Panwar, Pratyush Singh, Suman Saha, Danda~Pani Paudel, and Luc~Van Gool.
\newblock Unsupervised compound domain adaptation for face anti-spoofing.
\newblock In {\em {FG}}, pages 1--8. {IEEE}, 2021.

\bibitem{DBLP:journals/jmlr/Pearl10}
Judea Pearl.
\newblock Causal inference.
\newblock In {\em {NIPS} Causality: Objectives and Assessment}, volume~6 of
  {\em {JMLR} Proceedings}, pages 39--58. JMLR.org, 2010.

\bibitem{DBLP:conf/icb/PurnapatraSBDYM21}
Sandip Purnapatra, Nic Smalt, Keivan Bahmani, Priyanka Das, David Yambay, Amir
  Mohammadi, Anjith George, Thirimachos Bourlai, S{\'{e}}bastien Marcel,
  Stephanie Schuckers, Meiling Fang, Naser Damer, Fadi Boutros, Arjan Kuijper,
  Alperen Kantarci, Basar Demir, Zafer Yildiz, Zabi Ghafoory, Hasan Dertli,
  Hazim~Kemal Ekenel, Son Vu, Vassilis Christophides, Dashuang Liang, Guanghao
  Zhang, Zhanlong Hao, Junfu Liu, Yufeng Jin, Samo Liu, Samuel Huang, Salieri
  Kuei, Jag~Mohan Singh, and Raghavendra Ramachandra.
\newblock Face liveness detection competition (livdet-face) - 2021.
\newblock In {\em International {IEEE} Joint Conference on Biometrics, {IJCB}
  2021, Shenzhen, China, August 4-7, 2021}, pages 1--10. {IEEE}, 2021.

\bibitem{DBLP:conf/iccv/Rao0L021}
Yongming Rao, Guangyi Chen, Jiwen Lu, and Jie Zhou.
\newblock Counterfactual attention learning for fine-grained visual
  categorization and re-identification.
\newblock In {\em {ICCV}}, pages 1005--1014. {IEEE}, 2021.

\bibitem{DBLP:conf/cvpr/SahaXKGCPG20}
Suman Saha, Wenhao Xu, Menelaos Kanakis, Stamatios Georgoulis, Yuhua Chen,
  Danda~Pani Paudel, and Luc~Van Gool.
\newblock Domain agnostic feature learning for image and video based face
  anti-spoofing.
\newblock In {\em {CVPR} Workshops}, pages 3490--3499. {IEEE}, 2020.

\bibitem{DBLP:conf/cvpr/ShaoLLY19}
Rui Shao, Xiangyuan Lan, Jiawei Li, and Pong~C. Yuen.
\newblock Multi-adversarial discriminative deep domain generalization for face
  presentation attack detection.
\newblock In {\em {CVPR}}, pages 10023--10031. Computer Vision Foundation /
  {IEEE}, 2019.

\bibitem{DBLP:conf/aaai/ShaoLY20}
Rui Shao, Xiangyuan Lan, and Pong~C. Yuen.
\newblock Regularized fine-grained meta face anti-spoofing.
\newblock In {\em {AAAI}}, pages 11974--11981. {AAAI} Press, 2020.

\bibitem{DBLP:conf/ai4i/ShimizuAOMHK18}
Ryota Shimizu, Kosuke Asako, Hiroki Ojima, Shohei Morinaga, Mototsugu Hamada,
  and Tadahiro Kuroda.
\newblock Balanced mini-batch training for imbalanced image data classification
  with neural network.
\newblock In {\em {AI4I}}, pages 27--30. {IEEE}, 2018.

\bibitem{DBLP:conf/cvpr/TzengHSD17}
Eric Tzeng, Judy Hoffman, Kate Saenko, and Trevor Darrell.
\newblock Adversarial discriminative domain adaptation.
\newblock In {\em {CVPR}}, pages 2962--2971. {IEEE} Computer Society, 2017.

\bibitem{JMLR:v9:vandermaaten08a}
Laurens van~der Maaten and Geoffrey Hinton.
\newblock Visualizing data using t-sne.
\newblock {\em Journal of Machine Learning Research}, 9(86):2579--2605, 2008.

\bibitem{DBLP:conf/nips/Vapnik91}
Vladimir Vapnik.
\newblock Principles of risk minimization for learning theory.
\newblock In {\em {NIPS}}, pages 831--838. Morgan Kaufmann, 1991.

\bibitem{DBLP:conf/cvpr/WangLYL22}
Chien{-}Yi Wang, Yu{-}Ding Lu, Shang{-}Ta Yang, and Shang{-}Hong Lai.
\newblock Patchnet: {A} simple face anti-spoofing framework via fine-grained
  patch recognition.
\newblock In {\em {CVPR}}, pages 20249--20258. {IEEE}, 2022.

\bibitem{DBLP:conf/icb/WangHSC19}
Guoqing Wang, Hu Han, Shiguang Shan, and Xilin Chen.
\newblock Improving cross-database face presentation attack detection via
  adversarial domain adaptation.
\newblock In {\em {ICB}}, pages 1--8. {IEEE}, 2019.

\bibitem{DBLP:conf/cvpr/Wang0SC20}
Guoqing Wang, Hu Han, Shiguang Shan, and Xilin Chen.
\newblock Cross-domain face presentation attack detection via multi-domain
  disentangled representation learning.
\newblock In {\em {CVPR}}, pages 6677--6686. Computer Vision Foundation /
  {IEEE}, 2020.

\bibitem{DBLP:journals/tifs/WangHSC21}
Guoqing Wang, Hu Han, Shiguang Shan, and Xilin Chen.
\newblock Unsupervised adversarial domain adaptation for cross-domain face
  presentation attack detection.
\newblock {\em {IEEE} Trans. Inf. Forensics Secur.}, 16:56--69, 2021.

\bibitem{DBLP:journals/tbbis/WangWDG22}
Zhuo Wang, Qiangchang Wang, Weihong Deng, and Guodong Guo.
\newblock Face anti-spoofing using transformers with relation-aware mechanism.
\newblock {\em {IEEE} Trans. Biom. Behav. Identity Sci.}, 4(3):439--450, 2022.

\bibitem{Tyler2015}
Tyler~Vander Weele.
\newblock {Explanation in causal inference: Methods for mediation and
  interaction}.
\newblock {\em Oxford University Press}, 2015.

\bibitem{msu_mfs}
Di Wen, Hu Han, and Anil~K. Jain.
\newblock Face spoof detection with image distortion analysis.
\newblock {\em {IEEE} Trans. Inf. Forensics Secur.}, 10(4):746--761, 2015.

\bibitem{DBLP:journals/tifs/WenHJ15}
Di Wen, Hu Han, and Anil~K. Jain.
\newblock Face spoof detection with image distortion analysis.
\newblock {\em {IEEE} Trans. Inf. Forensics Secur.}, 10(4):746--761, 2015.

\bibitem{DBLP:conf/eccv/XuLNX14}
Zheng Xu, Wen Li, Li Niu, and Dong Xu.
\newblock Exploiting low-rank structure from latent domains for domain
  generalization.
\newblock In {\em {ECCV} {(3)}}, volume 8691 of {\em Lecture Notes in Computer
  Science}, pages 628--643. Springer, 2014.

\bibitem{DBLP:journals/tcsv/YanZH22}
Wenjun Yan, Ying Zeng, and Haifeng Hu.
\newblock Domain adversarial disentanglement network with cross-domain
  synthesis for generalized face anti-spoofing.
\newblock {\em {IEEE} Trans. Circuits Syst. Video Technol.}, 32(10):7033--7046,
  2022.

\bibitem{DBLP:journals/corr/YangLL14}
Jianwei Yang, Zhen Lei, and Stan~Z. Li.
\newblock Learn convolutional neural network for face anti-spoofing.
\newblock {\em CoRR}, abs/1408.5601, 2014.

\bibitem{DBLP:journals/tbbis/YuLSXZ21}
Zitong Yu, Xiaobai Li, Jingang Shi, Zhaoqiang Xia, and Guoying Zhao.
\newblock Revisiting pixel-wise supervision for face anti-spoofing.
\newblock {\em {IEEE} Trans. Biom. Behav. Identity Sci.}, 3(3):285--295, 2021.

\bibitem{DBLP:journals/pami/YuWQLLZ21}
Zitong Yu, Jun Wan, Yunxiao Qin, Xiaobai Li, Stan~Z. Li, and Guoying Zhao.
\newblock {NAS-FAS:} static-dynamic central difference network search for face
  anti-spoofing.
\newblock {\em {IEEE} Trans. Pattern Anal. Mach. Intell.}, 43(9):3005--3023,
  2021.

\bibitem{DBLP:journals/spl/ZhangZLQ16}
Kaipeng Zhang, Zhanpeng Zhang, Zhifeng Li, and Yu Qiao.
\newblock Joint face detection and alignment using multitask cascaded
  convolutional networks.
\newblock {\em {IEEE} Signal Proc. Lett.}, 23(10):1499--1503, 2016.

\bibitem{DBLP:conf/iccv/ZhangWWLKLG21}
Xiheng Zhang, Yongkang Wong, Xiaofei Wu, Juwei Lu, Mohan~S. Kankanhalli,
  Xiangdong Li, and Weidong Geng.
\newblock Learning causal representation for training cross-domain pose
  estimator via generative interventions.
\newblock In {\em {ICCV}}, pages 11250--11260. {IEEE}, 2021.

\bibitem{DBLP:conf/eccv/ZhangYLYYSL20}
Yuanhan Zhang, Zhenfei Yin, Yidong Li, Guojun Yin, Junjie Yan, Jing Shao, and
  Ziwei Liu.
\newblock Celeba-spoof: Large-scale face anti-spoofing dataset with rich
  annotations.
\newblock In {\em {ECCV}. Glasgow, UK, August 23-28, 2020,}, volume 12357 of
  {\em Lecture Notes in Computer Science}, pages 70--85. Springer, 2020.

\bibitem{casia_fas}
Zhiwei Zhang, Junjie Yan, Sifei Liu, Zhen Lei, Dong Yi, and Stan~Z. Li.
\newblock A face antispoofing database with diverse attacks.
\newblock In {\em {ICB}}, pages 26--31. {IEEE}, 2012.

\bibitem{DBLP:conf/iclr/ZhouY0X21}
Kaiyang Zhou, Yongxin Yang, Yu Qiao, and Tao Xiang.
\newblock Domain generalization with mixstyle.
\newblock In {\em {ICLR}}. OpenReview.net, 2021.

\end{thebibliography}
}

\end{document}